\documentclass{article}
\usepackage[a4paper,margin=1in]{geometry}
\usepackage{hyperref}
\usepackage{booktabs}
\usepackage{array} 
\usepackage{graphicx} 

\title{OpenAg: Democratizing Agricultural Intelligence}
\author{
  Srikanth Thudumu and Jason Fisher \\
  \textit{Institute of Applied Artificial Intelligence and Robotics (IAAIR)} \\
  Germantown, TN, USA, 38139 \\
  \textit{srikanth@iaair.ai; jason@iaair.ai}
}
\date{}
\begin{document}

\maketitle
\begin{abstract}
\noindent
Agriculture is undergoing a major transformation driven by artificial intelligence (AI), machine learning, and knowledge representation technologies. However, current agricultural intelligence systems often lack contextual understanding, explainability, and adaptability—especially for smallholder farmers with limited resources. General-purpose large language models (LLMs), while powerful, typically lack the domain-specific knowledge and contextual reasoning needed for practical decision support in farming. They tend to produce recommendations that are too generic or unrealistic for real-world applications. To address these challenges, we present OpenAg, a comprehensive framework designed to advance agricultural artificial general intelligence (AGI). OpenAg combines domain-specific foundation models, neural knowledge graphs, multi-agent reasoning, causal explainability, and adaptive transfer learning to deliver context-aware, explainable, and actionable insights. The system includes: (i) a unified agricultural knowledge base that integrates scientific literature, sensor data, and farmer-generated knowledge; (ii) a neural agricultural knowledge graph for structured reasoning and inference; (iii) an adaptive multi-agent reasoning system where AI agents specialize and collaborate across agricultural domains; and (iv) a causal transparency mechanism that ensures AI recommendations are interpretable, scientifically grounded, and aligned with real-world constraints. OpenAg aims to bridge the gap between scientific knowledge and the tacit expertise of experienced farmers to support scalable and locally relevant agricultural decision-making.

\subsubsection*{Keywords}
Adaptive Transfer Learning, AGI for Agriculture, Agricultural Artificial General Intelligence, Causal Explainability, Context-Aware Decision Support, Domain-Specific Foundation Models, Knowledge Graphs, Small Language Models, Multi-Agent Systems, Neural Knowledge Graphs, OpenAg, Smallholder Farming, Tacit Knowledge Integration.

\end{abstract}

\section{Introduction}

Agriculture stands at a critical juncture, facing unprecedented challenges of feeding a growing global population while confronting climate change, water scarcity, and environmental degradation \cite{mana2024sustainable}. The sector is undergoing a paradigm shift driven by artificial intelligence (AI), data science, and knowledge representation technologies. Modern precision agriculture generates vast quantities of data from diverse sources including remote sensing, IoT devices, farm machinery, weather stations, and expert knowledge. However, farmers, researchers, and policymakers struggle with fragmented, non-standardized, and often opaque AI-driven recommendations.

Current machine learning (ML) models applied to agriculture typically lack three critical capabilities: contextual understanding of complex agricultural systems, interpretability of decision rationales, and adaptability across diverse agricultural environments. These limitations severely restrict their real-world applicability, particularly in regions where historical and tacit knowledge play crucial roles in successful farming practices. While AI-powered decision support systems can provide real-time insights and personalized advice on farming methods and crop management, effectively implementing such systems remains challenging \cite{asolo2024ai}. Moreover, general large language models and world models lack specialized agricultural knowledge and contextual reasoning, often producing recommendations that are too generic or impractical, especially for smallholder farmers with resource constraints. These models also do not capture tacit knowledge—intuitive expertise developed through experience that is not easily articulated but is essential for decision-making. To this end, we identified several key challenges that impede the effective application of AI in agriculture. First, there is a fundamental lack of structured knowledge representation. Agricultural data is inherently heterogeneous, unstructured, and often region-specific, making it difficult to integrate into cohesive decision support systems. The Open Ag Data Alliance (OADA) has identified this problem as ``walled gardens of incompatible data'' that overwhelm farmers and limit the potential of data-driven agriculture \cite{oada_principles}. Second, existing AI-driven agricultural decision systems typically function as black boxes, providing recommendations without transparent reasoning processes. This absence of explainability limits trust and adoption among stakeholders. As Brugler et al. \cite{brugler2024improving} highlight, one major barrier to adoption of decision support systems is ``the hesitancy of farmers to change from their trusted advisor to a computer program that often behaves as a black box''. Third, most agricultural AI models suffer from limited adaptability, requiring large amounts of localized training data and lacking transferability across regions, crops, and farming practices. This constraint is particularly problematic in agricultural contexts, where conditions vary dramatically across geography, climate zones, and growing seasons.

To address these challenges, we propose the OpenAg initiative which includes a comprehensive framework designed to ingest, structure, and harmonize agricultural knowledge from multiple sources; enable real-time, multi-agent reasoning for AI-driven decision-making; leverage neural knowledge graphs for structured and scalable intelligence; ensure decision explainability through causal AI models; and adapt to new regions via transfer learning and continual learning mechanisms. The remainder of this paper is organized as follows: Section 2 discusses the background and significance of AI-driven agricultural intelligence. Section 3 details the proposed OpenAg architecture. Section 4 describes the implementation methodology. Section 5 presents our conclusions and future directions.

\section{Background and Significance}

\subsection{The Need for AI-Driven Agricultural Intelligence}

Traditional agricultural practices rely heavily on human expertise, historical data, and field observations, which are highly localized and difficult to scale. The global agricultural sector faces unprecedented challenges including climate change, water scarcity, soil degradation, and increasing food demand. These challenges necessitate more efficient, sustainable, and adaptive farming practices that can be supported by advanced AI systems. AI and machine learning have demonstrated tremendous potential to optimize agricultural decision-making through applications in crop monitoring, disease detection, yield prediction, and resource management. 
Ramcharan et al. \cite{ramcharan2017deep} showed that applying deep learning techniques for early disease classification allows farmers to take timely action, thereby helping to increase yield without inflicting unnecessary environmental damage through excessive use of fertilizers or pesticides. Farhan et al. \cite{farhan2023transfer} demonstrated that transfer learning techniques can dramatically improve crop type classification through remote sensing, with applications extending to crop yield estimation, disease detection, and crop phenotyping. However, existing AI solutions in agriculture face fundamental limitations. Data heterogeneity presents a significant challenge, as agricultural data comes in various forms including sensor readings, satellite imagery, textual reports, and expert insights. Integrating these diverse data types requires sophisticated data processing and knowledge representation techniques. Most current AI solutions in agriculture employ deep learning or complex statistical models that lack interpretability. This ``black box'' nature undermines trust and adoption, as farmers and agronomists need to understand the reasoning behind recommendations to properly apply their domain expertise. Additionally, models trained on specific regions or crops often perform poorly when applied to new contexts due to differences in soil, climate, farming practices, and other variables. 

Sustainable agriculture requires a more sophisticated approach to AI \cite{mana2024sustainable}that can overcome these limitations while providing actionable, trustworthy, and adaptable intelligence across diverse agricultural contexts.

\subsection{Knowledge Graphs in Agriculture}

Semantic knowledge graphs have emerged as powerful tools for representing complex, interconnected information in fields such as bioinformatics, finance, and healthcare \cite{galluzzo2024comprehensive}. These graph-based representations enable structured reasoning across diverse data points and relationships \cite{peng2023knowledge}. In the agricultural domain, knowledge graphs offer significant potential by providing a flexible, extensible framework for representing and reasoning with agricultural knowledge. Chhetri et al. \cite{chhetri2023towards} demonstrated that combining deep learning techniques with knowledge graphs could address critical challenges in agricultural AI systems, including the lack of explainability, the absence of contextual information, and the difficulty of incorporating domain expert knowledge. Their system for cassava root disease classification achieved 90.5\% prediction accuracy while generating user-level explanations about diseases and incorporating contextual and domain knowledge.

Despite these promising developments, the application of knowledge graphs in agriculture has been limited by several factors. The agricultural domain lacks standardized terminologies and conceptual frameworks across different regions and specialties, making it challenging to create unified knowledge representations. Additionally, agricultural systems involve intricate relationships between crops, soil, climate, pests, and management practices that are difficult to capture in traditional knowledge representations. Integrating diverse sources of agricultural knowledge—from scientific literature to farmer expertise—into a structured, machine-readable format requires sophisticated knowledge extraction and integration techniques.

OpenAg addresses these challenges by combining neural knowledge representation with multi-agent reasoning and causal explainability, creating a more comprehensive and adaptable framework for agricultural intelligence.

\subsection{Explainability and Causal AI in Decision Support}

For agricultural stakeholders including farmers, agronomists, and policymakers, AI recommendations must be not only accurate but also trustworthy and interpretable \cite{mallinger2024responsible, tzachor2022responsible, shams2024enhancing}. Users need to understand the rationale behind AI-driven recommendations to assess their validity, apply them appropriately, and integrate them with their own expertise. The significance of explainability in agricultural AI is multifaceted. Transparent reasoning processes help users trust AI recommendations, particularly in high-stakes decisions involving substantial investments or potential yield impacts. As agricultural policies increasingly incorporate sustainability metrics, transparent AI decision-making becomes essential for compliance and certification. Furthermore, explainable models can reveal novel insights about agricultural systems, contributing to scientific understanding and improved practices. Brugler et al. \cite{brugler2024improving} identified several barriers to adoption of decision support systems in precision agriculture, including the hesitancy of farmers to change from their trusted advisor to a computer program that often behaves as a ``black box''. Their research emphasizes that decision support systems with clear explanations are more likely to be adopted and properly utilized by farmers and agricultural professionals. Causal reasoning models offer a promising approach by enabling ``why'' explanations rather than just ``what'' predictions \cite{kiciman2023causal}. By modeling cause-effect relationships within agricultural systems, causal AI can provide more meaningful and actionable explanations for agricultural recommendations. OpenAg integrates causal reasoning, semantic transparency, and explainability as core components of its agricultural intelligence pipeline, addressing a critical gap in current agricultural AI systems.

\section{Proposed Architecture}

The OpenAg architecture is a multi-layered, AI-driven framework, designed to create a pathway toward Artificial General Intelligence (AGI) in agriculture by seamlessly integrating diverse data flows and advanced reasoning. It begins with multi-modal knowledge ingestion, aggregating technical, scientific, experiential, and tacit data into a unified agricultural knowledge base (see Fig 1). The OpenAg architecture consists of six primary components designed to work together to deliver context-aware, explainable, and adaptive agricultural intelligence. This section describes each component and its role within the overall architecture.

\begin{figure*}
    \centering
    \includegraphics[width=1.1\textwidth,height=1.2\textwidth]{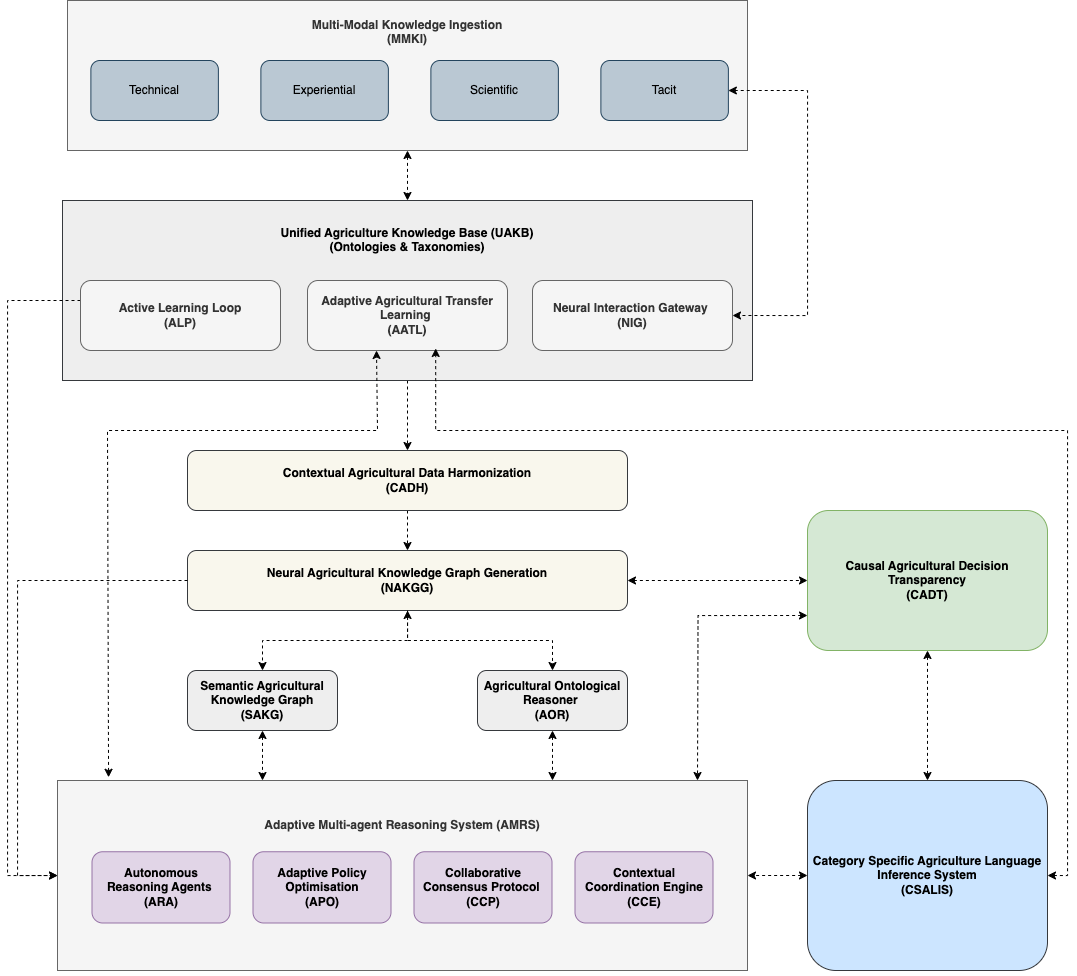}
    \caption{The Proposed Architecture of OpenAg}
    \label{fig:architecture}
\end{figure*}

\subsection{Multi-Modal Knowledge Ingestion}

The knowledge ingestion component serves as the entry point for diverse agricultural information into the OpenAg framework. It processes multiple data types from various sources, including technical reports and documentation, scientific papers, field observations, sensor data, expert interviews, remote sensing imagery, and historical management records. The ingestion process involves several key steps. First, ontology-based structuring maps incoming data to agricultural ontologies to standardize terminology and concepts. Named entity recognition then identifies key agricultural concepts, locations, crops, practices, and other relevant entities within the data. Relationship extraction uses natural language processing techniques to identify connections between entities, such as cause-effect relationships and correlational patterns. Data standardization normalizes numerical values to enable cross-source comparisons, while quality assessment procedures ensure data reliability. This approach builds upon previous work in agricultural data management, particularly the multi-agent copilot approach proposed by Pan et al. \cite{pan2024building}, while extending it with more sophisticated knowledge extraction capabilities. By transforming heterogeneous agricultural data into structured knowledge, this component establishes the foundation for subsequent reasoning and decision support processes.

\subsection{Unified Agricultural Knowledge Base (UAKB)}

The UAKB serves as the central repository for structured agricultural knowledge within OpenAg. It extends beyond traditional databases by incorporating semantic relationships and reasoning capabilities, enabling more sophisticated knowledge representation and retrieval. The UAKB is built on comprehensive agricultural ontologies that define concepts, properties, and relationships in the domain. It incorporates temporal dynamics through versioning and timestamping to track changes over time, critical for understanding seasonal patterns and long-term trends in agricultural systems. Spatial indexing enables location-specific reasoning, accounting for the inherent geographic variability of agricultural conditions. Additionally, the UAKB represents uncertainty explicitly, capturing confidence levels for knowledge assertions and enabling probabilistic reasoning under uncertainty. It also includes mechanisms for detecting and resolving contradictory information from different sources, a common challenge when integrating diverse agricultural knowledge. This component draws inspiration from previous work on knowledge graphs for agricultural applications while adding dynamic updating and reasoning capabilities. By creating a unified, structured representation of agricultural knowledge, the UAKB enables more sophisticated reasoning and decision support than would be possible with traditional database approaches.

\subsection{Neural Agricultural Knowledge Graph Engine}

The Neural Knowledge Graph Engine transforms the static knowledge in the UAKB into a dynamic, learnable representation that can support sophisticated reasoning and inference. This component leverages recent advances in graph neural networks and knowledge graph embedding techniques while adapting them to the specific challenges of agricultural knowledge representation. The engine employs Graph Neural Networks (GNNs) to learn representations of agricultural entities and relationships, enabling inductive reasoning across the knowledge graph. Entity linking connects new observations and inputs to existing knowledge entities, allowing the system to incorporate new information into its existing knowledge structure. Relational inference predicts missing relationships and properties within the knowledge graph, filling knowledge gaps based on existing patterns. Temporal reasoning models time-dependent processes such as crop growth cycles and seasonal patterns, while spatial reasoning accounts for geographic variation in agricultural conditions and practices. These capabilities are essential for agricultural decision support, as they enable the system to reason about dynamic agricultural processes across time and space. By combining neural network learning with structured knowledge representation, this component enables more flexible and adaptive reasoning than would be possible with either approach alone. The neural knowledge graph approach allows OpenAg to discover patterns and relationships that might not be explicitly encoded in the knowledge base, while maintaining the interpretability and structure of graph-based representations.

\subsection{Adaptive Multi-Agent Reasoning System (AMRS)}

The AMRS orchestrates multiple AI agents that collaborate to generate agricultural insights and recommendations. This multi-agent approach enables more specialized and flexible reasoning than would be possible with a monolithic system, while facilitating explainability and adaptability across diverse agricultural contexts. The system includes various agent types with specialized capabilities. Crop specialists focus on specific crops and their requirements, while resource managers optimize water, fertilizer, and other resource allocation. Risk assessors evaluate potential threats from weather, pests, and market conditions, and sustainability monitors ensure recommendations align with environmental and social sustainability goals. Integration coordinators harmonize insights from other agents into cohesive recommendations. Agents collaborate through consensus protocols that enable them to negotiate and reach agreement on recommendations. Hierarchical planning breaks complex decisions into manageable sub-tasks, while knowledge sharing allows agents to exchange insights and learned patterns. Meta-reasoning at the system level evaluates agent performance and reliability, enabling the system to improve over time. This approach builds upon previous multi-agent systems in agriculture while adding more sophisticated reasoning capabilities and explicit focus on collaboration and consensus. By distributing agricultural reasoning across specialized agents, the AMRS can address complex agricultural decisions that require diverse expertise and perspective.

\subsection{Causal Agricultural Decision Transparency (CADET)}

CADET ensures that all recommendations from OpenAg are transparent and explainable to users, addressing a critical barrier to adoption of agricultural decision support systems. This component employs several techniques to provide meaningful explanations for agricultural recommendations. Causal inference models identify and represent cause-effect relationships in agricultural systems, moving beyond correlational patterns to provide more meaningful explanations. Counterfactual reasoning enables ``what-if'' scenarios that help users understand why certain recommendations are made over alternatives. Narrative generation produces natural language explanations tailored to user expertise, translating complex AI reasoning into accessible explanations. Visual explainability creates intuitive visualizations of decision rationales, helping users understand complex relationships and trade-offs. Uncertainty communication clearly conveys confidence levels and potential risks, enabling users to make informed decisions under uncertainty. This component addresses the critical need for explainability in agricultural AI systems, as highlighted by recent research on barriers to adoption of decision support systems \cite{brugler2024improving}. By providing transparent, causal explanations for agricultural recommendations, CADET enhances user trust and enables more effective collaboration between human expertise and AI-driven insights.

\subsection{Adaptive Agricultural Transfer Learning}

This component enables OpenAg to adapt its knowledge and recommendations to new agricultural contexts with minimal additional training, addressing the challenge of data scarcity in many agricultural settings. It employs several advanced techniques to enable knowledge transfer across regions, crops, and farming systems. Meta-learning approaches train the system to learn efficiently across different agricultural scenarios, enabling rapid adaptation to new contexts. Domain adaptation techniques adjust models to account for differences between source and target domains, such as variations in climate, soil, and farming practices. Few-shot learning generates accurate predictions with limited examples from new contexts, while continual learning updates models incrementally without catastrophic forgetting of previously acquired knowledge. Knowledge distillation transfers insights from complex models to more deployable, lightweight versions, enabling deployment in resource-constrained environments. This approach is particularly valuable in agricultural contexts, where computational resources may be limited and deployment conditions may vary widely. This component builds upon recent advances in transfer learning for crop classification \cite{farhan2023transfer} while extending them to the broader context of agricultural decision support. By enabling efficient knowledge transfer across diverse agricultural contexts, the Adaptive Agricultural Transfer Learning component enhances OpenAg's applicability and value across global agriculture.

\section{Proposed Methodology}

The implementation of OpenAg requires sophisticated methodological approaches for each component. This section details the key methodological elements for realizing the OpenAg framework.

\subsection{Knowledge Acquisition and Structuring}

The process of acquiring and structuring agricultural knowledge involves several key methodological steps. First, we develop comprehensive agricultural ontologies that extend existing resources with detailed representations of crop phenology, soil properties, weather patterns, pest and disease lifecycles, farm management practices, resource utilization metrics, and sustainability indicators. This ontological framework provides the foundation for structured knowledge representation throughout the system.

For multi-modal data processing, we employ specialized techniques for each data type. Text processing uses advanced natural language processing models fine-tuned on agricultural corpora to extract entities, relationships, and facts from scientific literature and technical reports. Sensor data processing applies time-series analysis and anomaly detection to IoT data from fields and equipment. Image processing uses computer vision to analyze satellite imagery and field photos, while expert knowledge elicitation captures tacit knowledge through structured interviews and knowledge engineering techniques.

Knowledge integration combines these diverse sources through entity resolution, contradiction detection, confidence scoring, semantic alignment, and temporal contextualization. This integrated approach enables OpenAg to create a comprehensive, structured representation of agricultural knowledge that can support sophisticated reasoning and decision support.

\subsection{Neural Knowledge Graph Implementation}

The Neural Knowledge Graph Engine employs advanced techniques for knowledge representation and reasoning. We implement a hybrid GNN architecture combining Graph Convolutional Networks \cite{corso2024graph} for capturing local neighborhood structures, Graph Attention Networks for differentially weighting relationships, and Temporal Graph Networks \cite{rossi2020temporal} for modeling dynamic processes. These models are trained on both supervised tasks (e.g., link prediction) and self-supervised tasks (e.g., context prediction) to learn rich representations of agricultural entities and relationships.

Knowledge embedding techniques transform the graph into a continuous vector space, enabling efficient similarity computation and pattern discovery. We employ methods such as TransE \cite{bordes2013transe}, TransE-MTP \cite{li2024transe} for modeling simple relational patterns, RotatE \cite{sun2019rotate} for capturing complex relation patterns, and RGCN \cite{schlichtkrull2018modeling} for handling heterogeneous graphs with multiple relation types. These embeddings allow the system to perform efficient similarity searches, analogy reasoning, and pattern discovery across the agricultural knowledge space.

The system combines multiple reasoning mechanisms, including deductive reasoning (applying logical rules), inductive reasoning (generalizing from examples), abductive reasoning (inferring likely explanations), and analogical reasoning (applying knowledge from similar situations). These approaches are coordinated through a meta-reasoning layer that selects appropriate strategies based on the problem and available knowledge.

\subsection{Multi-Agent System Design}

The AMRS is implemented using a cooperative multi-agent reinforcement learning framework \cite{li2019cooperative}. Each agent has a three-tier architecture consisting of a perception layer that processes inputs, a reasoning layer that applies domain-specific expertise, and a communication layer that exchanges information with other agents. Agents employ a combination of rule-based reasoning, Bayesian networks, and neural models depending on their specific domain.

Agents coordinate through a consensus protocol that includes proposal generation, consistency checking, utility evaluation, consensus formation, and explanation generation \cite{amirkhani2022consensus, qin2016recent, du2024survey}. This approach enables collaborative decision-making while maintaining transparency and accountability. The system improves over time through reinforcement learning (from recommendation outcomes), imitation learning (from expert demonstrations), curriculum learning (tackling increasingly complex scenarios), and federated learning (sharing patterns while maintaining privacy).

\subsection{Causal Explainability Implementation}

The CADET component implements causal explainability \cite{carloni2025role} through several sophisticated techniques. Causal discovery combines constraint-based methods (e.g., PC and FCI algorithms), score-based methods, and domain knowledge integration to identify causal relationships in agricultural systems. Causal inference then applies techniques such as do-calculus, propensity score matching, instrumental variables, and Causal Bayesian Networks to reason about interventions and effects.

Explanations are generated through a multi-step process: relevance determination identifies key causal factors, counterfactual generation creates alternative scenarios, narrative construction produces coherent natural language explanations, visualization creation develops intuitive visual representations, and uncertainty communication expresses confidence levels and risks. This approach builds upon work by Chhetri et al. \cite{chhetri2023towards} on user-level explainability in agricultural systems, extending it with more sophisticated causal modeling.

\subsection{Transfer Learning Methodology}

OpenAg employs several transfer learning techniques \cite{pan2009survey, bozinovski2020reminder} to enable adaptation across different agricultural contexts. Domain adaptation methods such as domain-adversarial training, gradient reversal, and MMD minimization reduce the discrepancy between source and target domains. Meta-learning approaches including model-agnostic meta-learning (MAML) enable rapid adaptation to new crops, regions, or farming practices with minimal data.

Continual learning techniques prevent catastrophic forgetting when adapting to new contexts. We implement elastic weight consolidation to preserve important parameters, experience replay to maintain examples from previous contexts, knowledge distillation to transfer knowledge between models, and modular architecture to isolate context-specific components while sharing general knowledge. This comprehensive approach to transfer learning extends recent work on agricultural applications \cite{farhan2023transfer} to create a more adaptable and widely applicable decision support framework.

\section{Conclusion}

This paper has presented OpenAg, a novel framework for context-aware, explainable, and adaptive agricultural intelligence. By integrating multi-agent reasoning, neural knowledge graphs, causal explainability, and transfer learning, OpenAg addresses critical challenges in applying AI to agricultural decision support that have limited the effectiveness and adoption of existing approaches. The proposed architecture transforms heterogeneous fragmented agricultural knowledge into structured, machine-readable intelligence through a modular system featuring a Unified Agricultural Knowledge Base, Neural Agricultural Knowledge Graph Engine, Adaptive Multi-Agent Reasoning System, and Causal Agricultural Decision Transparency mechanism. This integrated approach enables more sophisticated reasoning and decision support than would be possible with any single technique. OpenAg's multi-agent reasoning system orchestrates specialized AI agents that collaborate to generate agricultural insights and recommendations, enabling more flexible and comprehensive decision support than monolithic systems. The neural knowledge graph approach combines the strengths of structured knowledge representation with the pattern recognition capabilities of neural networks, creating a more powerful and adaptable knowledge foundation. Causal explainability addresses a critical barrier to adoption by providing transparent, meaningful explanations for agricultural recommendations, building trust and enabling effective human-AI collaboration. The transfer learning methodology enables OpenAg to adapt to diverse agricultural contexts with minimal additional training, addressing the challenge of data scarcity in many agricultural settings. This capability is particularly important for deploying advanced agricultural intelligence in regions where data may be limited but the need for resource-efficient agricultural productivity is pressing.

Future work on OpenAg will focus on enhancing several key aspects of the framework. First, we will develop more automated knowledge acquisition methods to streamline the integration of new agricultural knowledge. Second, we will create lightweight versions of OpenAg components for edge deployment in resource-constrained environments. Third, we will establish longitudinal studies to evaluate the long-term effects of OpenAg recommendations on agricultural sustainability and productivity. Additional research directions include cultural adaptation of the system to diverse contexts, improved uncertainty modeling and communication, enhanced human-AI co-learning, and integration with agricultural policy frameworks. Through these efforts, we aim to create a more accessible, effective, and widely applicable agricultural intelligence framework that can support sustainable food production worldwide. OpenAg represents a significant step toward democratizing advanced agricultural intelligence, enabling farmers, researchers, and policymakers to benefit from AI-driven insights while maintaining transparency, adaptability, and contextual relevance. As agriculture faces unprecedented challenges from resource constraints, land degradation, and growing food demand, initiatives such as OpenAg will play an increasingly important role in supporting sustainable, productive farming practices across diverse agricultural landscapes.

\section*{Acknowledgments}

We thank Richard Lackey, Chairman of the World Food Bank, and Joel Cunningham, CEO of the Hamara Group, for their valuable feedback and continued support of this work. Their insights and encouragement have helped shape the direction and impact of this research.

\bibliographystyle{IEEEtran}
\bibliography{references}

\end{document}